\documentclass{article} % For LaTeX2e
\usepackage{iclr2027_conference,times}

\usepackage{amsmath,amsfonts,bm}

\def\eqref#1{equation~\ref{#1}}
\def\1{\bm{1}}

\DeclareMathAlphabet{\mathsfit}{\encodingdefault}{\sfdefault}{m}{sl}
\SetMathAlphabet{\mathsfit}{bold}{\encodingdefault}{\sfdefault}{bx}{n}

\usepackage{hyperref}
\usepackage{url}
\usepackage{xspace}
\usepackage{booktabs}
\usepackage{graphicx}
\usepackage{pifont}
\usepackage{multirow}
\usepackage[dvipsnames]{xcolor}
\usepackage{tcolorbox} 
\tcbuselibrary{listings,breakable}
\usepackage{enumitem}

\newcommand{\cmark}{\ding{51}}
\newcommand{\xmark}{\ding{55}}

\newcommand{\method}{\textsc{RECAP}\xspace}

\definecolor{darkgreen}{RGB}{0,100,0}
\definecolor{darkred}{RGB}{139,0,0}

\title{
\method: Training Long-horizon LLM Agents with Context Compaction
}

\author{
\makebox[\dimexpr\textwidth-2\tabcolsep\relax][c]{%
\normalfont
\begin{tabular}[t]{@{}c@{}}
\textbf{Yulun Jiang}$^{1,2}$ \quad
\textbf{Maria Brbic}$^{1,\dagger}$ \quad
\textbf{Michael Moor}$^{2,\dagger}$ \\[6pt]
$^{1}$EPFL \quad $^{2}$ETH Zurich
\end{tabular}
}
}

\hypersetup{colorlinks=true,linkcolor=blue!60!black,citecolor=blue!60!black,urlcolor=blue!60!black}
\iclrfinalcopy % Uncomment for camera-ready version, but NOT for submission.
\begin{document}

\maketitle
\begingroup
\renewcommand{\thefootnote}{\fnsymbol{footnote}}
\footnotetext[2]{Equal advising. Correspondence to \texttt{mbrbic@epfl.ch} and \texttt{michael.moor@bsse.ethz.ch}. Code is available at \texttt{\color{blue!50!black} \hyperlink{https://github.com/eth-medical-ai-lab/recap}{https://github.com/eth-medical-ai-lab/recap}}.}
% \hyperlink{https://github.com/mlbio-epfl/recap}{https://github.com/mlbio-epfl/recap}}\michael{Yulun, this is not updated yet.}.} 
\endgroup

\begin{abstract}

Context compaction enables large language model agents to tackle long-horizon tasks within limited context budgets by replacing interaction histories with compressed summaries.
However, learning effective compaction jointly with the agent policy remains challenging, as summaries affect task success over many subsequent interactions.
We introduce \method, a reinforcement learning (RL) framework based on the principle that a summary should preserve the behavior induced by the original interaction history.
We formalize this principle with a behavior distortion objective that measures the divergence between the agent’s subsequent behavior under compressed and original contexts. 
We show that the gradient of the distortion objective decomposes into two complementary learning signals: distortion-based reward for summary generation and self-distillation for subsequent actions. 
Experiments on Qwen3.5-4B show that \method consistently outperforms the task-reward-only RL baselines on both environments, leading to an absolute improvement of $4.6\%$ and $4.3\%$ on BrowseComp-plus and SWE-bench Verified, respectively.
Ablations further highlight the value of jointly supervising summary generation and post-compaction behavior for \method. 
% Our code is publicly available at {\color{Cerulean} \hyperlink{https://github.com/mlbio-epfl/recap}{https://github.com/mlbio-epfl/recap}}.

\end{abstract}

\section{Introduction}

% LLM Agent background, RL \& post-training. Long-horizon tasks

% Context management and memory. E.g., compaction, subagent, retrieval, recursive... Context compaction is the most basic and widely used. Question: how to train this behavior.

% RL training is challenging. Existing attempts ... However, ...

% Our motivation, intuition, objective, analysis, integration ...

% Experiment results ...

Large language model (LLM) agents have demonstrated strong capabilities in solving complex tasks through reasoning, tool use, and multi-step interactions with their environments~\citep{yao2022react,wei2022chain,nakano2021webgpt,smith2025-openhands-context-condensensation}.
Recent advances in post-training, particularly reinforcement learning (RL), have further improved agent performance in various challenging domains such as deep-research~\citep{jin2025search, li2025webthinker}, software engineering~\citep{wei2025swe,yang2026swe} and scientific discovery ~\citep{huang2026autonomous,falck2026training}.
However, as task horizons increase, LLM agents accumulate growing interaction histories that can exceed the context limit before task
completion~\citep{beltagy2020longformer,liu2024lost}.
Thus, scaling LLM agents to long-horizon tasks requires effective management of the information retained in the context for
subsequent decisions.

% Context compaction provides a natural solution to this challenge.
% When approaching the limit, the agent is prompted to generate a compressed summary to replace the current interaction history.
% These summaries carry forward task progress and relevant observations while reducing the amount of history retained in context. 
% Compaction thus enables longer interactions, but its effectiveness depends on both what the summary preserves and how the agent acts from the resulting context.
% Then the agent will continue working with the compressed context, allowing it to work under fixed context budget.
% Due to its flexibility, it has been adopted in various agent harness.

Context compaction provides a straightforward approach to overcome context limits by consolidating an overlong interaction history into a compressed summary~\citep{zhou2026mem1,yu2026memagent}.
This simple yet effective mechanism allows agents to operate within a fixed context budget and has been widely adopted in various agent harnesses~\citep{smith2025-openhands-context-condensensation}.
In particular, earlier works have explored context compaction through carefully designed summarization prompts~\citep{kang2025acon} or separately optimized compressors~\citep{yi2026adacom}.
More recent works focused on jointly training the agent policy and compaction through RL~\citep{lu2025supo,sun2025foldgrpo,li2026compactionrl}.
However, these approaches typically rely on final task rewards to optimize both compaction and task execution.
In long trajectories, the effect of a compaction decision might propagate through many subsequent actions or later compactions, making credit assignment difficult. This raises a fundamental question: \textit{Is there a more direct learning signal for context compaction that can guide training long-horizon LLMs beyond task success?}

In this paper, we introduce \method (\textbf{RE}inforcement learning for \textbf{C}ontext compaction and \textbf{A}gent \textbf{P}olicy optimization), a framework for training long-horizon agents with context compaction.
Our key insight is that a useful summary should retain the information needed for subsequent decisions, thereby \textit{preserving the behavior induced by the original interaction history}.
We formalize this principle with a behavior distortion objective that measures the token-level KL divergence between the agent's responses under compressed and original contexts.
Here, the same policy conditioned on the original context serves as a self-teacher, specifying the behavior to be preserved after compaction.

% Here, the same policy conditioned on the original context serves as a self-teacher, providing a behavioral reference for the agent after compaction.

% We formulate the behavior distortion objective as the token-level KL divergence over the actions after compaction.
% And the full-history policy therefore serves as a self-teacher specifying the behavior that should be preserved after compaction.

We decompose the gradient of the distortion objective and show that it provides two complementary learning signals: \textit{(i)} distortion-based reward for summary generation, and \textit{(ii)} self-distillation for actions after compaction.
The distortion-based reward assigns higher credit to summaries that induce less behavior distortion compared to the original context. 
The context self-distillation directly trains the subsequent actions using the full-context policy as the self teacher, providing dense token-level supervision to the same policy conditioned on the compressed context~\citep{hubotter2026reinforcement, zhao2026self}.
We combine these two learning signals with the task-success objective, jointly optimizing context compaction and the agent policy within a unified RL framework, which does not require any additional rollouts or environment interactions.

% Our analysis shows that this objective provides two complementary learning signals for summary generation and subsequent actions.
% For summary generation, negative distortion serves as a scalar reward, favoring summaries that better preserve the agent's subsequent behavior.
% Meanwhile, actions generated under the compressed context receive dense token-level supervision from the self-teacher through self-distillation~\citep{hubotter2026reinforcement,zhao2026self}.
% We integrate these signals with the task-success objective to jointly optimize context compaction and the agent policy within a unified RL framework.

We evaluate \method on two representative long-horizon environments, BrowseComp-plus~\citep{chen2025browsecomp} and SWE-bench Verified~\citep{jimenez2024swe}, covering deep-research and software engineering tasks. 
Using Qwen3.5-4B with a 32K working context limit, \method 
outperforms the task-reward-only RL baselines by $4.6\%$ absolute score on the BrowseComp-plus and $4.3\%$ on the SWE-bench Verified, respectively.
These gains are accompanied by fewer tool calls and lower token usage in the trajectory, demonstrating improved efficiency of the trained agents.
Moreover, we find the advantage of \method is more evident under a tighter 16K working-context limit, suggesting that behavior-preserving supervision is particularly useful in the scenario where the agent frequently compact its history to solve the task.

\begin{figure}
    \centering
    \includegraphics[width=0.98\linewidth]{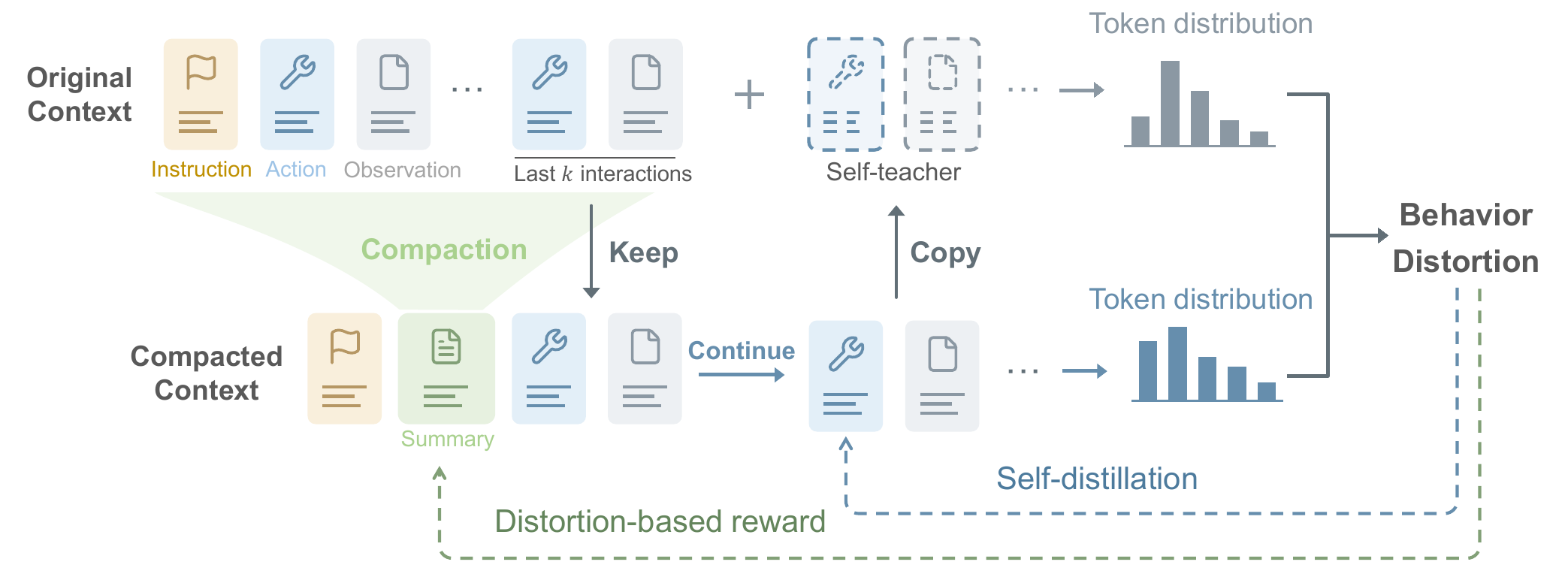}
    \vspace{-0.5em}
    \caption{\textbf{Overview of the behavior distortion objective in \method.} We measure the effect of context compaction through the KL divergence between the subsequent actions conditioned on original and compacted contexts, which results in a distortion-based reward for summary generation (green) and self-distillation for post-compaction actions (blue).
    }
    \label{fig:method}
    \vspace{-1em}
    \end{figure}

\section{Related Work}

\textbf{RL for LLM agent}.
Reinforcement learning (RL) has become a central component of LLM post-training, improving instruction following and reasoning through human feedback and verifiable rewards~\citep{ouyang2022training,shao2024deepseekmath,schulman2017proximal,guo2025deepseek,yu2025dapo}.
Recent work extends RL to multi-turn agents~\citep{chen2025loop, wang2025ragen,wang2026reinforcement,jiang2026metarl}, enabling models to learn search and software-engineering behaviors through interactions with tools and environments~\citep{nakano2021webgpt,jin2025search,zheng2025deepresearcher,jain2025r2e,wei2025swe}.
As task horizons increase, sparse and delayed rewards make it difficult to assign credit to individual actions. Recent work addresses this challenge through step-level advantage estimation and turn-level reward assignment, providing finer-grained supervision for agent training~\citep{feng2025group, wang2025spa,tan2026hindsight,dong2026agentic,tao2026trace}.
Complementing these efforts, \method augments task rewards with a behavior distortion objective that provides localized learning signals for both summary generation and post-compaction actions.

\textbf{On-policy self-distillation}.
On-policy distillation trains language models using teacher predictions on student-generated trajectories, providing a dense token-level supervision~\citep{agarwal2024policy,lu2025onpolicydistillation,ma2026mopd,ye2026policy}. 
Recent self-distillation methods construct a teacher from the same model by conditioning it on privileged information, such as verified solutions or environment feedback~\citep{hubotter2026reinforcement,zhao2026self,kim2026does,yang2026self,shenfeld2026self}. This approach has also been extended to multi-turn agents, where skill-conditioned self-distillation complements task-reward optimization~\citep{lu2026sdar,wang2026skill}.
Our work uses the original interaction history as privileged context to supervise the agent's post-compaction behavior.
We further use the resulting behavioral discrepancy to assign credit to summaries, jointly optimizing context compaction and the agent policy.

% Alongside reward-based optimization, on-policy distillation provides dense token-level supervision by matching a teacher’s predictions on student-generated trajectories~\citep{agarwal2024policy,lu2025onpolicydistillation,ma2026mopd,ye2026policy}.
% Recent works of self-distillation use the model conditioned on privileged information as a teacher to supervise the same model~\citep{hubotter2026reinforcement,zhao2026self,kim2026does,yang2026self,shenfeld2026self}.
% Integration to agentic tasks~\cite{lu2026sdar,wang2026skill}
% Our work complements task-reward-based RL with contextual self-distillation, using the original interaction history as privileged information to supervise the agent's behavior after context compaction.

% RL for Agentic tasks, towards more complex problems such as deepsearch~\citep{nakano2021webgpt,jin2025search,zheng2025deepresearcher}, swe~\cite{jain2025r2e,yang2026swe,wei2026swe}, web~\citep{nakano2021webgpt}...
% Recent advanced of OPD~\citep{lu2025onpolicydistillation, agarwal2024policy,ma2026mopd} and self-distillation~\cite{hubotter2026reinforcement,zhao2026self,kim2026does,yang2026self}

\textbf{Context management for long-horizon task}.
Existing approaches manage growing interaction histories through context compaction~\citep{wu2025resum,kang2025acon,ye2026agentfold,yuan2026memsearcher}, selective memory retention and retrieval~\citep{packer2023memgpt,xie2026echo,zhang2026memory}, or context decomposition across sub-agents and recursive calls~\citep{zhang2024chain,zhang2025recursive}.
Recent work uses end-to-end RL to jointly optimize context management and task execution~\citep{lu2025supo,sun2025foldgrpo,li2026compactionrl}.
For context compaction, however, task outcomes provide only indirect supervision for summaries whose effects unfold over subsequent interactions.
\method complements task-success rewards with a behavior distortion objective that supervises both summary generation and post-compaction actions.

% Three categoris:
% Summary and compaction~\citep{zhou2026mem1,yu2026memagent}
% Active retireval~\cite{xie2026echo,li2026acm}
% Recursive~\cite{zhang2025recursive,zhang2024chain}

% Attempts for RL training on long-horizon tasks~\cite{li2026compactionrl,lu2025supo,sun2025foldgrpo,wu2025resum}
\section{Background}

In this section, we formalize the agent rollouts with context compaction and provide a preliminary for key concepts we build upon.

\textbf{Standard agent loop.} Let $I$ denote the system prompt and task instruction. A standard LLM-agent trajectory can be represented by:
\begin{equation}
\tau = (I,a_1, o_1, a_2, o_2, \ldots,a_T, o_T),
\end{equation}
where $a_t$ is the model response and $o_t$ is the environment observation at step $t \in \{1\ldots T\}$. The context available for LLM agent before action $a_t$ is
\begin{equation}
h_t=(I,a_1, o_1, a_2, o_2, \ldots,a_{t-1}, o_{t-1}).
\end{equation}
At each step, the agent generates a response from its policy $a_t \sim \pi_\theta(\cdot \mid h_t)$, after which the environment produces an observation through the transition function $o_t \sim P(\cdot \mid h_t, a_t)$. The new action-observation pair $(a_t, o_t)$ is directly appended to the context for the next turn.
Together, this process describes the standard agent loop under the case of sufficient context window.

\textbf{Context compaction}. 
% For long-horizon tasks, the context accumulates through the process of reasoning and environment interactions. To prevent overlong context that exceeds the context limit of LLM, one natural choice is to compact the context when the context length exceeds a predefined threshold $L_\text{trigger}$. 
For overlong trajectory, we compact the accumulated context whenever its length exceeds a fixed budget.
Specifically, when $|h_t| \geq L_\text{trigger}$ at step $t$ with $L_\text{trigger}$ being the fixed budget, the agent is prompted to generate a compressed summary $m_t$ of the interaction history
\begin{equation}
    m_t \sim \pi_\theta(\cdot \mid h_t \oplus q_\mathrm{comp}),
\end{equation}
where $q_\text{comp}$ is the fixed compression instruction and $\oplus$ denotes the prompt concatenation. The compaction instructs the agent to create a handoff summary on the current progress, key observation and potential next step. The content of the $q_\text{comp}$ is provided in Appendix~\ref{app:prompts}. 

After that, the compressed context is constructed as:
\begin{equation}
\tilde h_t = (I, m_t, \underbrace{a_{t-k}, o_{t-k}, \ldots, a_{t-1}, o_{t-1}}_{\text{Last $k$ interactions}})
\label{eq:tilde_h_t}
\end{equation}
where the task instruction and the last $k$ steps are preserved, and the intermediate interactions are replaced by the generated summary $m_t$.
The agent continues to work on task with the compressed context until the task end or the next compaction.

A long-horizon task might contain multiple compactions at steps $t_1<\cdots<t_K$, which naturally split the entire trajectory into multiple segments:
\begin{equation}
    \tilde \tau = (I, \sigma_0, m_{t_1}, \sigma_{1}, \ldots, m_{t_K}, \sigma_K),
\end{equation}
where each segment $\sigma_k$ is the collection of action and observation pairs in sequence.
At a later compaction, the agent again summarizes its current working context, including the previous summary and subsequent interactions, and applies the same rule to reconstruct the new context.

\textbf{RL training with compaction}. Training an agent with learned context compaction involves optimizing both the task policy and the generation of compression summaries. 
Recent works typically adopt end-to-end reinforcement learning (RL) to jointly optimize task actions and compression summaries~\citep{lu2025supo,li2026compactionrl}, using task success as the training objective: 
\begin{equation}
J_{\mathrm{task}}(\theta)=\mathbb{E}_{\tilde\tau\sim \pi_\theta}\left[R(\tilde\tau)\right],
\label{eq:J_task}
\end{equation}
where $R(\tilde\tau)\in \{0,1 \}$ denotes the binary reward of task-success and is assigned to both regular action steps and compaction steps. The policy can be optimized with standard policy-gradient methods such as PPO~\citep{schulman2017proximal} or GRPO~\citep{shao2024deepseekmath}.

However, the final task reward only evaluates trajectory as a whole, providing only indirect supervision for context compaction via task success. The difficulty is that in a long-horizon trajectory the effect of an individual compaction depends on many subsequent actions and potentially further compactions. %, making its contribution to the final outcome
%difficult to identify. 
As a result, the trajectory-level reward provides a noisy credit assignment for compaction decisions and obscures the effect of any single compaction. This limitation  was also reported in recent works ~\citep{sun2025foldgrpo,xie2026echo,yi2026adacom}. Motivated by this, we propose to use a complementary learning signal that evaluates the effect of each compaction directly.
 % on long-horizon context and memory management

\section{RECAP}

We introduce \method, an RL framework for jointly optimizing long-horizon agent policies and context compaction. Rather than supervising compaction solely through the final task-success signal, we provide a more direct learning signal based on its effect on the agent's subsequent behavior. Our central idea is that a good compaction should preserve the behavior induced by the original interaction history. Based on this principle, we introduce a \textit{behavior distortion objective} that in addition to the task success provides learning signal for agent training. We first formulate the objective, then derive the corresponding updates for summaries and continuations, and finally integrate them into joint policy optimization.

% We introduce \method, an RL framework for jointly optimizing long-horizon agent policies and context compaction. Rather than supervising compaction solely through the final task-success signal, we provide a more direct learning signal based on its effect on the agent's subsequent behavior. Our central idea is that a good compaction should preserve the behavior induced by the original interaction history. Based on this principle, we introduce a behavior distortion objective that directly trains both the compaction and the post-compaction policy. We first formulate this objective and then describe how it is integrated into end-to-end policy optimization.

\textbf{For clarity}. Throughout this section, we present the distortion objective only for a single compaction event in the trajectory. Our formulation naturally extends to the scenarios with multiple compactions, which we defer to Appendix~\ref{app:multiple_compaction}.

\subsection{Behavior Distortion Objective}
In addition to the final task success, we introduce a behavior distortion objective for jointly training the agent policy and context compaction. The key intuition is that \textit{the context compaction should preserve the agent's behavior in subsequent interactions}. In other words, after a compaction at step $t$, the behavior induced by the compressed context $\tilde h_t$ should remain close to that induced by the original context $h_t$. At the policy level, this discrepancy can be measured by the KL divergence:
\begin{equation}
D_{\mathrm{KL}}
\left(
\pi_{\theta}(\cdot \mid \tilde h_t)
\; \middle\| \;
\pi_{\theta}(\cdot \mid h_t)
\right).
\end{equation}
We measure the distortion through the subsequent interaction segment in the trajectory. Specifically, let $\sigma_t^+ = (a_t, o_t, \ldots,a_T,o_T)$ denote the action--observation segment after the compaction at step $t$, ending at the next compaction or at trajectory termination. For a generated summary $m_t$, we define its behavior distortion as:
\begin{equation}
D_t(m_t)= 
\mathbb{E}_{
\sigma_t^+ \sim \pi_\theta(\cdot \mid \tilde h_t) 
}
\left[
\sum_{i\in\sigma_t^+}
\sum_{j=1}^{|a_i|}
D_{\mathrm{KL}}
\left(
\pi_\theta(\cdot \mid \tilde h_t, s_{i,j})
\;\middle\|\;
\mathrm{sg}\left[\pi_\theta(\cdot \mid h_t, s_{i,j}) \right]
\right)
\right],
\label{eq:D_t}
\end{equation}
where $s_{i,j}$ is the shared action--observation prefix preceding $j$-th token of action $a_i$ within $\sigma_t^+$,
and $\mathrm{sg}(\cdot)$ denotes the stop-gradient operator.
Thus, $D_t(m_t)$ measures the accumulated token-level KL divergence over the action tokens in the segment $\sigma_t^+$.
The two policies share the same prefix $s_{i,j}$ and differ only in whether they're conditioned on  $\tilde h_t$ or $h_t$.
The full-history policy therefore serves as a self-teacher specifying the behavior that should be preserved after compaction.

% For long-horizon agent, a compaction should induce minimal behavioral distortion over the subsequent interaction segment. We therefore use the negative distortion as a direct learning signal. For a compaction occurring at step $t$, we define:
We define the compaction objective by minimizing this behavioral
distortion:
\begin{equation}
J_\mathrm{comp} (\theta)
= 
\mathbb{E}_{m_t \sim \pi_\theta(\cdot \mid h_t \oplus q_\mathrm{comp})} 
\left[ - D_t(m_t) \right].
\label{eq:J_comp}
\end{equation}
Maximizing $J_{\mathrm{comp}} (\theta)$ therefore encourages the compressed-context behavior to remain close to the corresponding full-context behavior. 

Together with the final task success in Eq.(\ref{eq:J_task}), the \method's final objective is:
\begin{equation}
J_\mathrm{total}(\theta)=J_\mathrm{task}(\theta)+\lambda J_\mathrm{comp}(\theta),
\end{equation}
where $\lambda$ controls the relative strength of behavior preservation and task success. 
% Intuitively, $J_\mathrm{comp}(\theta)$ regularizes the behavioral change induced by compaction, while $J_\mathrm{task}(\theta)$ encourages the agent to retain behaviors that are useful for solving the task.
Here $J_\mathrm{comp}(\theta)$ serves as an auxiliary regularizer for behavior distortion under context compaction, while $J_\mathrm{task}(\theta)$ encourages behaviors that are useful for solving the task.

% In summary, the distortion objective provides a local learning signal around each compaction event complementary to the task success. Importantly, this signal depends on both the summary that constructs the compressed context and the policy behavior that follows from it. We next show that these two dependencies naturally induce two complementary gradient terms.

\subsection{Learning Summaries and Continuations}

Optimizing the behavior distortion objective in Eq.(\ref{eq:J_comp}) affects the policy through two pathways:
\begin{enumerate}[label=\textbf{(\Roman*)}]
\item The compression summary $m_t \sim \pi_\theta(\cdot \mid h_t \oplus q_\mathrm{comp})$.
\item The post-compression action-observation segments $\sigma_t^+ \sim \pi_\theta(\cdot \mid \tilde h_t)$.
\end{enumerate}
These two pathways are reflected in the gradient of the compaction objective $J_\mathrm{comp}$. Conditional on the fixed pre-compaction history $h_t$, the gradient of $J_\mathrm{comp}$ in Eq.(\ref{eq:J_comp}) can be decomposed as
\begin{align} \label{eq-decomp}
\nabla_\theta J_\mathrm{comp}(\theta) 
% &= \nabla_\theta \mathbb{E}_{m_t} \left[-D_t (m_t) \right]  \nonumber \\ 
&= 
\underbrace{
\mathbb{E}_{m_t}
\left[
-D_t(m_t) \nabla_\theta \mathrm{log} \pi_\theta(m_t \mid h_t \oplus q_\mathrm{comp})
\right]
}
_{g_\mathrm{sum}}
+
\underbrace{
\mathbb{E}_{m_t}
\left[
-\nabla_\theta D_t(m_t)
\right]
}
_{g_\mathrm{post}},
\end{align}
where $g_\mathrm{sum}$ corresponds to the gradient assigned to the generated summary and $g_\mathrm{post}$ corresponds to the gradient of the post-compression segments. 

% \textbf{Learning the compaction policy}. The summary $m_t$ is a stochastic action sampled by the policy. Its effect on subsequent behavior is therefore optimized through policy gradient:
% \begin{equation}
% g_\mathrm{sum} = \mathbb{E}_{m_t}
% \left[
% -D_t(m_t) \nabla_\theta \mathrm{log} \pi_\theta(m_t \mid h_t \oplus q_\mathrm{comp})
% \right].
% \end{equation}
% where the distortion $-D_t(m_t)$ serves as an direct reward assigned to the summary content $m_t$. Intuitively, summaries that introduce less behavioral distortion receive higher credit.

\textbf{Learning the summary from distortion-based reward}.
The first term $g_\mathrm{sum}$ corresponds to a standard policy-gradient update for summary generation, where the negative behavior distortion serves as the reward signal. For each generated summary $m_t$, we sample a single subsequent segment $\sigma_t^+$ to estimate the distortion, denoted by $\widehat D_t$. The scalar reward $-\widehat D_t$ is then assigned to all tokens in $m_t$, giving higher credit to summaries that induce less distortion in the agent's subsequent behavior.

\textbf{Learning the continuation through self-distillation}.
The second term $g_\mathrm{post}$ provides supervision to the agent's actions after compaction. 
We treat the policy conditioned on the full context $\pi_\theta(\cdot \mid h_t)$ as the teacher and the policy conditioned on the compressed context $\pi_\theta(\cdot \mid \tilde h_t )$ as the student.
% The continuation is generated only by the student. 
% For each generated token, the teacher evaluates the same token under the same post-compaction prefix $s_{i,j}$, differing only in whether it conditions on the original context $h_t$ or the compressed context $\tilde h_t$.
Specifically, the student generates the continuation $\sigma_t^+$, and then the teacher evaluates each token under the same post-compaction prefix $s_{i,j}$, differing only in the context choice of $h_t$ or $\tilde h_t$.
We therefore optimize this term using an on-policy self-distillation estimator, which provides dense token-level supervision throughout the continuation:
\begin{equation}
% \nabla_\theta D_t(m_t)
g_\mathrm{post}^\mathrm{SD}
= 
\mathbb{E}_{m_t, \sigma_t^+}
\left[ 
\sum_{i\in \sigma_t^+} 
\sum_{j=1}^{|a_i|} 
\mathrm{log} 
\frac{\pi_\theta(a_{i,j} \mid h_t, s_{i,j})}{\pi_\theta (a_{i,j} \mid \tilde h_t, s_{i,j})} 
\cdot 
\nabla_\theta \mathrm{log} 
\pi_\theta(a_{i,j} \mid \tilde h_t, s_{i,j}) 
\right],
\end{equation}
where the expectation follows the summary and continuation sampling described above. 
% In practice, we reuse the continuation sampled to estimate $\widehat D_t$, so the teacher only scores the student's trajectory and no additional rollout is required. 
We provide the derivation in Appendix~\ref{app:derivation}. 
Therefore, for each token in the continuation, the self-distillation objective provides token-level advantage as 
$
A^{\mathrm{SD}}_{i,j}
=
\log\frac{
\pi_\theta(a_{i,j}\mid h_t,s_{i,j})
}{
\pi_\theta(a_{i,j}\mid \tilde h_t,s_{i,j})
}.
$
This token-level advantage can then be combined directly with the summary-level distortion-based reward and the task advantage in the joint policy update.

Both terms can be expressed as policy-gradient updates with token-level weights. We summarize the learning signal induced by the compaction objective as the compaction advantage:
\begin{equation}
A^{\mathrm{comp}}
=
\begin{cases}
-\widehat D_t-b_t,
& \text{for each summary token in}\ m_t, \\
A^{\mathrm{SD}}_{i,j},
& \text{for each action token}\ a_{i,j}\ \text{in}\ \sigma_t^+ ,
\end{cases}
\end{equation}
where $b_t$ is an optional baseline for the negative-distortion reward. Action tokens outside the post-compaction continuation receive no compaction advantage. Thus, the summary receives a shared distortion-based signal, while the continuation receives token-level self-distillation signals.

\subsection{Joint Optimization with Task Reward}
The compaction advantage can be directly combined with the task advantage used by the underlying RL algorithm. Let $A^{\mathrm{task}}$ denote the task advantage. We obtain the joint advantage as:
\begin{equation}
A^\mathrm{total} = A^{\mathrm{task}} + \lambda A^{\mathrm{comp}},
\end{equation}
where the corresponding token-level values are used for each summary or action token. In this way, the task reward supervises the entire rollout, while the behavior distortion objective provides additional supervision to the summary and the actions that follow it.

We optimize $A^{\mathrm{total}}$ using the same policy-gradient objective as the task-only RL baseline of GRPO~\citep{shao2024deepseekmath,lu2025supo}. 
Details of the training process are provided in Appendix~\ref{app:algorithm_implementation}.
Importantly, \method requires \textit{no additional rollouts or environment interactions}, as both the distortion-based reward and self-distillation signal are computed from the same sampled trajectory. The overall training time of \method remains roughly on par with the RL baselines.

% \subsection{Joint Optimization with Task Reward}

% The three learning signals—task success, summary-level distortion, and post-compaction distortion—can be optimized within a common policy-gradient framework. Let \(A_i^{\mathrm{task}}\) denote the advantage obtained from the terminal task reward. For each trainable policy token, we construct the combined advantage:
% \begin{equation}
% A_i = A_i^{\mathrm{task}} + \lambda A_i^{\mathrm{comp}},
% \end{equation}
% where $A_i^{\mathrm{comp}}$ takes the corresponding summary-level or post-compaction advantage when applicable and is zero otherwise.

% Given trajectories collected under a behavior policy \(\pi_{\mathrm{old}}\), we define the token-level importance ratio
% \begin{equation}
%    r_i(\theta) = \frac{ \pi_\theta(y_i\mid c_i) }{ \pi_{\mathrm{old}}(y_i\mid c_i) }. 
% \end{equation}
% We optimize the combined signal using the standard clipped policy-gradient surrogate,
% \begin{equation}
% J_{\mathrm{sur}}(\theta) = \mathbb E \left[ \sum_i \min \left( r_i(\theta)A_i,\, \operatorname{clip} \left( r_i(\theta), 1-\epsilon, 1+\epsilon \right)A_i \right) \right].
% \end{equation}
% Equation specifies the underlying learning objective of our method, while Eq. (14) is the practical surrogate used to optimize its policy-gradient estimators.

% Finally we get a good training algorithm, integrated with reward for summary and self-distillation for post-compaction summary.

% It does not require any extra rollout process.

% Everything can be jointly optimized end-to-end with policy gradient

\section{Experiment}

\subsection{Experimental setting}

% Environment: Broswecomp-plus for deepsearch agents and SWE-bench Verified for SWE agent. setting for each environment

% Training details: Model. Context setting. Training parameters. Baselines.

\textbf{Datasets}. We evaluate our method on two representative long-horizon agent tasks: deep search and software engineering.
For deep search, we use BrowseComp-Plus~\citep{chen2025browsecomp}, with 300 questions for training and 150 held-out questions for evaluation.
The provided tools are \texttt{search(query)}, \texttt{get\_document(docid)} and \texttt{finish(answer)}.
For software engineering, we select 300 tasks from SWE-smith~\citep{yang2026swe} for training and evaluate on all 500 tasks in SWE-bench Verified~\citep{jimenez2024swe}.
The available tools are \texttt{bash(command)}, \texttt{str\_replace\_editor(command, path)} and \texttt{finish()}.
In both environments, agents interact with the environment through the provided tools until task termination, with context compaction triggered when the context reaches the specified budget.
All methods use the same training data and configurations within each environment.
Details of data selection and environment prompts are provided in Appendix~\ref{app:experiment_details} and~\ref{app:prompts}.

\textbf{Training Details.}
We conduct our main experiments with Qwen3.5-4B~\citep{qwen35blog}, with thinking mode disabled. For both environments, we use a 32K working-context limit and allow up to three compaction events per trajectory. After each compaction, the compressed context retains the most recent interaction ($k=1$).
We allow up to $100$ interaction steps for BrowseComp-Plus and $200$ for SWE-agent, and use the same compaction prompt across both environments.
We additionally evaluate a 128K context setting without compaction as a reference.
For \method, we set $\lambda$ based on a held out validation set with 100 examples for BrowseComp-Plus and SWE-agent, respectively.
RL training is implemented in a fully asynchronous setting using the VeRL framework~\citep{sheng2024hybridflow}.
Across all the training runs, we use a rollout batch size of 16, a group size of 8, a learning rate of $1\times10^{-6}$, and train for 150 steps.
During evaluation, we sample responses with temperature of $1.0$, top-$p$ of $0.95$, and top-$k$ of $20$.
We report accuracy averaged over four evaluation runs.
Full training and evaluation configurations are provided in Appendix~\ref{app:experiment_details}. 

\textbf{Baselines.} We compare \method to task-reward-only GRPO~\citep{shao2024deepseekmath} with the same 32K working-context limit and compaction configuration as \method.
Following the end-to-end formulation of \citet{lu2025supo}, this method jointly optimizes the agent policy and compaction summaries using the final task-success rewards.
For reference, we also report results for GRPO trained and evaluated with a 128K context window without compaction.
We additionally report results from other open models using a 128K context window.

\subsection{\method performance results}

The results on BrowseComp-plus and SWE-bench Verified are presented in Table~\ref{tab:main_results}. 
Under a 32K working-context limit, \method outperforms the GRPO baselines with compaction by $4.6\%$ and $4.3\%$ in accuracy on the BrowseComp-Plus and SWE-bench Verified, respectively.
Moreover, \method substantially closes the gap to RL training with a 128K context window.
For example, on BrowseComp-Plus, \method reaches $62.8\%$ accuracy, only $1.0\%$ below the 128K GRPO result, while using only one quarter of the peak context length.
On the SWE-bench Verified, \method achieves $51.1\%$ under context compaction, approaching the $55.5\%$ performance of the 128K setting.
These consistent improvements across both environments support the benefit of augmenting task-success optimization with behavior-distortion supervision when jointly training the agent policy and context compaction.

\begin{table}[t]
    \centering
    \vspace{-1em}
    \caption{
    \textbf{Main results with and without context compaction.}
     Peak length denotes the maximum working-context length. Accuracy is averaged over 4 runs. For Qwen3.5-4B, we report mean accuracy ± standard deviation across the evaluation runs.
    }
    \label{tab:main_results}
    \vspace{0.4em}

    \setlength{\tabcolsep}{3pt}
    \renewcommand{\arraystretch}{1.15}
    
    \begin{tabular*}{\linewidth}{
        @{\extracolsep{\fill}}lcccc@{}
    }
        \toprule
        \multirow{2}{*}{\textbf{Method}}
        & \multirow{2}{*}{
            \shortstack{\textbf{Peak Length}}
          }
        & \multirow{2}{*}{
            \shortstack{\textbf{Compaction}}
          }
        & {\textbf{BrowseComp-Plus}}
        & {\textbf{SWE-bench Verified}} \\
        & &
        & Acc. (\%) & Acc. (\%)  \\
        \midrule

        \multicolumn{5}{@{}l}{\textit{Open-model references}} \\
        Gemma-4-E4B  & 128K & \color{darkred}{\xmark} & 24.7 & 15.4 \\
        Qwen3-Coder-30B-A3B  & 128K & \color{darkred}{\xmark} & 24.7 & 43.7 \\
        Seed-OSS-36B & 128K & \color{darkred}{\xmark} & 29.3 & 50.0 \\
        % GPT-OSS-120B & 128K & \xmark & -- & -- \\
        GLM-4.5-Air  & 128K & \color{darkred}{\xmark} & 44.7 & 52.8 \\
        Qwen3.5-9B   & 128K & \color{darkred}{\xmark} & 53.0 & 56.8 \\
        % DeekSeek-V3.1  & 128K & \xmark & -- & -- \\
        \midrule

        \multicolumn{5}{@{}l}{
            \textit{Without context compaction}
        } \\
        \addlinespace[2pt]
        Qwen3.5-4B & 128K & \color{darkred}{\xmark} & 49.2 \footnotesize {{$\pm$ 0.8}} 
        % & 49.5 \footnotesize {{$\pm$ 1.4}} 
        & 49.8 \footnotesize {{$\pm$ 2.1}} 
        \\
        \ +\ GRPO & 128K & \color{darkred}{\xmark} & 63.8 \footnotesize {{$\pm$ 3.0}} 
        % & 53.1 \footnotesize {{$\pm$ 0.5}} 
        & 55.5 \footnotesize {{$\pm$ 1.2}} 
        \\ 
        \midrule

        \multicolumn{5}{@{}l}{
            \textit{With context compaction}
        } \\
        \addlinespace[2pt]
        Qwen3.5-4B & 32K & \color{darkgreen}{\cmark} & 40.7 \footnotesize {{$\pm$ 3.1}} 
        % & 43.9 \footnotesize {{$\pm$ 1.0}} 
        & 45.8 \footnotesize {{$\pm$ 1.3}} 
        \\
        \ +\ GRPO & 32K &  \color{darkgreen}{\cmark} & 58.2 \footnotesize {{$\pm$ 1.5}}  
        % & 44.8 \footnotesize {{$\pm$ 1.2}} 
        & 46.8 \footnotesize {{$\pm$ 1.4}} 
        \\
        \ + \textbf{\method (ours)} & 32K &  \color{darkgreen}{\cmark} & \bf 62.8 \footnotesize {{$\pm$ 3.9}}  
        % & \bf 49.6 \footnotesize {{$\pm$ 0.7}} 
        & \bf 51.1 \footnotesize {{$\pm$ 1.0}} 
        \\
        \bottomrule
    \end{tabular*}

\label{tab:main_results}
\end{table}

\begin{figure}[h]
    \centering
    \includegraphics[width=0.95\linewidth]{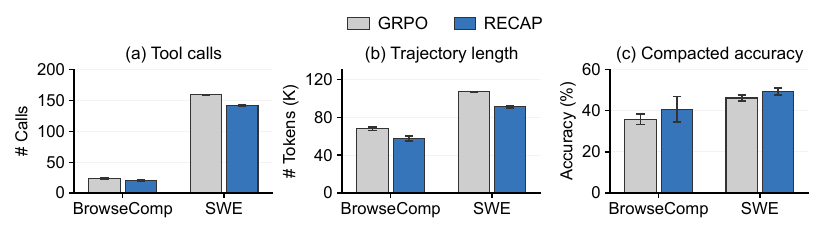}
    \vspace{-1em}
    \caption{\textbf{Interaction steps and accuracy across environments.} Comparison of (a) tool calls per trajectory, (b) cumulative tokens per trajectory, and (c) accuracy on trajectories with at least one compaction.
    \method uses fewer tool calls and tokens on average while achieving higher accuracy on compacted trajectories.
    }
    \label{fig:efficiency}
    \vspace{-0.5em}
\end{figure}

Beyond final task accuracy, we further compare the interaction steps and tokens per trajectory in the trained model in Figure~\ref{fig:efficiency} (a) and (b). Across both environments, \method uses substantially fewer tool calls and cumulative tokens per task compared to the RL baseline, demonstrating more efficient task execution.
We hypothesize that this may result from more effective behavior after context compaction, reducing redundant or repeated interactions.
Figure~\ref{fig:efficiency}(c) further reports accuracy on trajectories that trigger at least one compaction. 
\method consistently achieves higher accuracy on these trajectories, with a particularly pronounced improvement on the BrowseComp-Plus.
Together, these results suggest that the benefits of \method extend beyond overall task accuracy, improving both interaction efficiency and robustness following context compaction.

% We show the tool-call, response length, ect for two environments, 

% Analysis: Response length, summary quality, ect

\subsection{Ablation of Learning Signals}

We ablate the two learning signals introduced by the behavior distortion objective in Table~\ref{tab:ablation}.
We first assess the individual contribution of terms $g_{\mathrm{sum}}$ and $g_{\mathrm{post}}$ in Eq. (\ref{eq-decomp}) by augmenting the GRPO baseline with each term individually.
Adding $g_{\mathrm{sum}}$ term alone provides limited gains, with accuracy  decreasing by $1.5\%$ on the  BrowseComp-Plus while increasing by $1.1\%$ on the SWE-bench Verified.
In addition, $g_\mathrm{sum}$ alone also leads to more tool calls in both environments and produces substantially longer summaries on the BrowseComp-Plus.
In contrast, adding term $g_{\mathrm{post}}$ alone consistently improves task performance, reaching $60.3\%$ on BrowseComp-Plus and $47.1\%$ on SWE-bench Verified.
The trained model also uses fewer tool calls and shorter summaries in both environments.
We hypothesize that this is due to the self-distillation formulation of $g_{\mathrm{post}}$, which uses the full-context policy as a self-teacher to supervise post-compaction actions. Similar efficiency gains from self-distillation have also been reported in recent works~\citep{hubotter2026reinforcement,kim2026does}.

Importantly, the \method objective that combines both $g_{\mathrm{sum}}$ and $g_{\mathrm{post}}$ learning signals yields the best performance in both environments, reaching $62.8\%$ on BrowseComp-Plus and $51.1\%$ on SWE-bench Verified. 
This indicates that the two signals are complementary: $g_{\mathrm{post}}$ regularizes post-compaction behavior, while $g_{\mathrm{sum}}$ assigns credit to the summary that induces it. Jointly they provide more effective supervision for jointly learning context compaction and the agent policy.

\begin{table}[h]
    \centering
    \vspace{-0.5em}
    \caption{\textbf{Ablation of the two learning signals in \method}. We evaluate the individual and combined effects of terms $g_{\mathrm{sum}}$ and $g_{\mathrm{post}}$ in the training. We report accuracy, the average number of tool calls per trajectory, and the average summary length per compaction.
    }
    \label{tab:ablation}

    \vspace{1em}
    \setlength{\tabcolsep}{2.5pt}
    \renewcommand{\arraystretch}{1.15}

    \begin{tabular*}{\linewidth}{
        @{\extracolsep{\fill}}cccccccc@{}
    }
        \toprule

        \multicolumn{2}{c}{\bf Learning Signals}
        % & \multirow{2}{*}{$g_{\mathrm{sum}}$}
        % & \multirow{2}{*}{$g_{\mathrm{post}}$}
        & \multicolumn{3}{c}{\textbf{BrowseComp-Plus}}
        & \multicolumn{3}{c}{\textbf{SWE-bench Verified}} \\
        
        \cmidrule(lr){1-2}
        \cmidrule(lr){3-5}
        \cmidrule(l){6-8}

        $g_{\mathrm{sum}}$
        & $g_{\mathrm{post}}$
        & Acc. (\%)
        & \# Tools
        & Sum. Length
        & Acc. (\%)
        & \# Tools
        & Sum. Length \\

        \midrule

        \color{darkred}{\xmark}
        & \color{darkred}{\xmark}
        & 58.2
        & 23.4
        & 1146
        & 46.8
        & 158.9
        & 966 \\

        \color{darkgreen}{\cmark}
        & \color{darkred}{\xmark}
        & 56.7
        & 23.5
        & 1788
        & 47.9
        & 180.3
        & 917 \\

        \color{darkred}{\xmark}
        & \color{darkgreen}{\cmark}
        & 60.3
        & 21.2
        & 468
        & 47.1
        & 142.3
        & 557 \\

        \color{darkgreen}{\cmark}
        & \color{darkgreen}{\cmark}
        & \textbf{62.8}
        & 19.9
        & 709
        & \textbf{51.1}
        & 145.6
        & 1003 \\

        \bottomrule
    \end{tabular*}
\end{table}

\subsection{Generalization Across Context Windows}

\begin{figure}[h]
    \centering
    \includegraphics[width=0.95\linewidth]{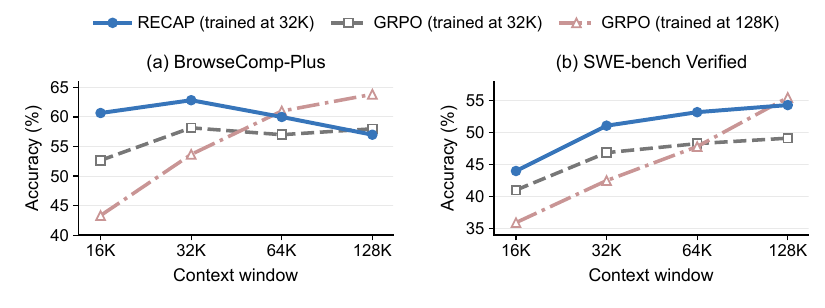}
    \vspace{-1.5em}
    \caption{\textbf{Performance across context windows}. Evaluation of trained agents on BrowseComp-Plus (left) and SWE-bench Verified (right) with an 128K interaction budget, using 16K $\times$ 8, 32K $\times$ 4, or 64K $\times$ 2 context segments, or 128K without compaction.
    }
    \label{fig:context_window}
\end{figure}

We further evaluate how the trained policies generalize across different context windows. We train RECAP and the GRPO baseline at 32K context window with compaction. 
For both, we fix the total interaction budget to 128K tokens and then vary the working-context window from 16K to 128K during evaluation. 
This setup results in increasingly frequent compaction when solving the long-horizon task as the context window becomes shorter. We further include the results of the GRPO trained at 128K context window without any compaction as a reference.
In both environments, \method obtains strong performance across different context windows compared to the GRPO baseline trained at the same context window (Figure~\ref{fig:context_window}). The gap is particularly large on shorter contexts where more frequent compaction is needed to solve the long-horizon task. Specifically, on the BrowseComp-Plus, \method maintains 60--63\% accuracy from 16K to 64K, substantially outperforming both GRPO variants at the shortest 16K window.
Similarly, on the SWE-bench Verified, \method outperforms GRPO trained at 32K across all context windows and also surpasses GRPO trained at 128K in all settings that require compaction.
In comparison, GRPO trained at 128K performs well in the full context setting but degrades considerably with shorter context windows that require compaction to solve the task.

% Across both environments, \method trained at 32K maintains strong performance beyond its training context window, with particularly clear advantages under shorter contexts that require frequent compaction.
% On BrowseComp-Plus, \method maintains around 60--63\% accuracy from 16K to 64K, substantially outperforming both GRPO variants at the shortest 16K window.
% On SWE-bench Verified, \method outperforms GRPO trained at 32K across all context windows and also surpasses GRPO trained at 128K in all settings that require compaction.
% In contrast, GRPO trained at 128K performs best when the full context is available but degrades considerably as the context window is shortened.
% Overall, these results suggest that explicitly training agents with context compaction improves their ability to operate under different context budgets, especially when frequent compaction is required.

\subsection{Ablation on Relative Weight}
\begin{figure}[h]
    \centering
    \vspace{-0.5em}
    \includegraphics[width=0.95\linewidth]{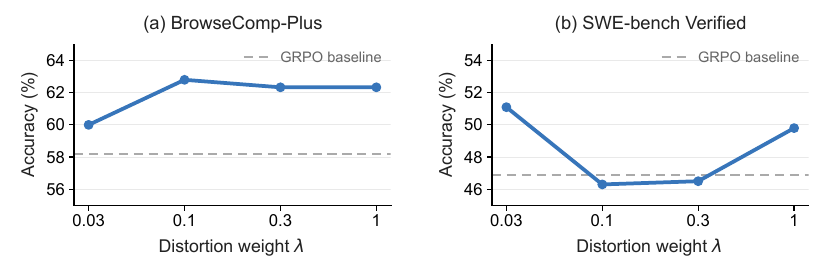}
    \vspace{-1em}
    \caption{\textbf{Ablation of the distortion objective weight $\lambda$.}
    Results with different weight of the behavior distortion objective on BrowseComp-Plus (left) and SWE-bench Verified (right).}
    \label{fig:lambda_ablation}
\end{figure}

We study the effect of the regularizer $\lambda$ in the behavior distortion objective in \method (Figure~\ref{fig:lambda_ablation}).
On the BrowseComp-Plus, performance consistently improves as $\lambda$ increases and remains stable for larger $\lambda$ values with no further gains. With all tested values of the regularizer $\lambda$, \method substantially outperforms GRPO baseline. On the SWE-bench Verified, performance is more sensitive to the relative weight. The effect of $\lambda$ is less monotonic, with $\lambda=0.03$ achieving the highest accuracy of 51.1\%.
We hypothesize that the difference is due to the higher complexity and longer interaction horizon on the tasks of SWE-bench Verified.
As shown in Figure~\ref{fig:efficiency} and Table~\ref{tab:ablation}, the trajectories in SWE-bench Verified involve substantially more interactions than BrowseComp-Plus, which could make it more challenging for the agent to preserve task-relevant information through compaction. In all experiments we set $\lambda$ based on the independent validation set which is detailed in Table~\ref{tab:app_lambda_selection} in Appendix~\ref{app:experiment_details}.

\section{Conclusion}
% In this paper, we introduced \method, a method to jointly learn long-horizon agent policies and compaction. TODO: summarize key findings and how this changes the literature - as well as limitations

In this paper, we introduced \method, an RL framework for jointly learning context compaction and long-horizon agent policies.
Its behavior distortion objective augments task-success optimization with two complementary learning signals: a distortion-based reward for summary generation and self-distillation for post-compaction actions.
Experiments on two representative long-horizon benchmarks, BrowseComp-Plus and SWE-bench Verified, demonstrate improved task performance and interaction efficiency over task-reward-only RL.
Our experiments focus on Qwen3.5-4B across two task domains, leaving broader evaluation across model scales, task domains, and longer interaction horizons to future work.
Real-world tasks might involve diverse information densities and representations, posing additional challenges for effective context compaction over long interaction horizons.
This also motivates exploring the benefits of \method alongside other non-trivial context management strategies, such as active forgetting and retrieval or task decomposition across recursive sub-agents.
It would also be interesting to integrate \method with critic-based RL methods such as PPO to improve credit assignment and sample efficiency in long-horizon training.
Overall, our findings suggest that the effects of compaction on subsequent behavior provides a good regularization signal that complements the task reward for training long-horizon agents under limited context budgets.

\section*{Acknowledgements}
We thank Ashraya Indrakanti, Siba Panigrahi, and Shuo Wen for valuable discussions and feedback throughout the project. 
We also thank the E2B for Research program for providing sandbox infrastructure that supported our large-scale RL experiments.
This project was supported by Swiss AI Initiative (projects \#a0174 and \#ab023).
M.B. gratefully acknowledges the support of the Swiss National Science Foundation (SNSF) starting grant TMSGI2\_226252/1, SNSF grant IC00I0\_231922, SNSF grant 10.004.411, Swiss Cancer League (KFS-6340-02-2025-R), Kamprad Foundation (Cancer Immunotherapy Networking Program) and the Swiss AI Initiative (\#ab006, \#ab023). 
M.B. is a CIFAR Fellow in the Multiscale Human Program. 
% \michael{I'm not sure it makes sense to acknowledge more SwissAI projects than actually were used here. I think it would be more appropriate to say "This project was supported by Swiss AI projects A and B (e.g. the small grant and the recent large Continualign).}

\bibliography{iclr2027_conference}
\bibliographystyle{iclr2027_conference}

\appendix
\clearpage
\section*{Appendix}

\section{Derivation of \method Objective}
\label{app:derivation}

We consider a single compaction event at a fixed history $h_t$.
The policy samples a summary on the current history $m_t \sim \pi_\theta(\cdot \mid h_t \oplus q_{\mathrm{comp}})$, constructs the compressed context $\tilde h_t$, and then continue to work on the compressed context with continuation $\sigma_t^+ \sim \pi_\theta(\cdot \mid \tilde h_t)$.
The continuation ends at the next compaction or task termination.
For each action token, let
\begin{equation}
d_{i,j}(\theta)
= D_{\mathrm{KL}}\!\left(
\pi_\theta(\cdot \mid \tilde h_t,s_{i,j})
\;\middle\|\;
\mathrm{sg}[\pi_\theta(\cdot \mid h_t,s_{i,j})]
\right),
\end{equation}
where $s_{i,j}$ is the shared continuation prefix and $\mathrm{sg}(\cdot)$ denotes the stop-gradient operator.
Writing $\sum_{i,j}$ for the sum over action tokens in $\sigma_t^+$, the behavior distortion objective is
\begin{equation}
J_{\mathrm{comp}}(\theta)
= \mathbb{E}_{m_t}[ - D_t(m_t)],
\qquad
D_t(m_t)
= \mathbb{E}_{\sigma_t^+ \mid m_t}\!\left[\sum_{i,j}d_{i,j}(\theta)\right].
\end{equation}
All expectations below follow these sampling distributions, with $h_t$ held fixed.

\subsection{Summary Gradient}

Applying the product rule and the log-derivative trick $\nabla_\theta \pi_\theta = \pi_\theta\nabla_\theta\log\pi_\theta$ gives the gradient decomposition in the main text:
\begin{align}
\nabla_\theta J_{\mathrm{comp}}(\theta)
&= \nabla_\theta\mathbb{E}_{m_t}\!\left[-D_t(m_t)\right]
\nonumber\\
&= -\nabla_\theta\sum_{m_t}
\pi_\theta(m_t\mid h_t\oplus q_{\mathrm{comp}})D_t(m_t)
\nonumber\\
&= -\sum_{m_t}\Bigl[
D_t(m_t)\nabla_\theta\pi_\theta(m_t\mid h_t\oplus q_{\mathrm{comp}})
\nonumber\\[-0.3em]
&\hspace{6em}
+\pi_\theta(m_t\mid h_t\oplus q_{\mathrm{comp}})\nabla_\theta D_t(m_t)
\Bigr]
\nonumber\\
&= \underbrace{\mathbb{E}_{m_t}\!\left[
-D_t(m_t)\nabla_\theta\log\pi_\theta(m_t\mid h_t\oplus q_{\mathrm{comp}})
\right]}_{g_{\mathrm{sum}}}
\nonumber\\
&\quad+\underbrace{\mathbb{E}_{m_t}\!\left[-\nabla_\theta D_t(m_t)\right]}_{g_{\mathrm{post}}}.
\label{eq:app_gradient_decomposition}
\end{align}
The first term $g_\mathrm{sum}$ is the policy gradient for summary generation, with negative distortion as its scalar reward.
In practice, we estimate $D_t(m_t)$ from one sampled continuation and assign the reward $-\widehat D_t$ to every summary token generated by the policy.

\subsection{Post-Compaction Self-Distillation}
For the second term $g_\mathrm{post}$, we hold the sampled summary fixed and expand the derivative through the continuation expectation:
\begin{align}
g_{\mathrm{post}}
&= -\mathbb{E}_{m_t}\!\left[
\nabla_\theta\mathbb{E}_{\sigma_t^+\mid m_t}\!\left[\sum_{i,j}d_{i,j}(\theta)\right]
\right]
\nonumber\\
&= -\mathbb{E}_{m_t,\sigma_t^+}\!\left[
\left(\sum_{i,j}d_{i,j}(\theta)\right)
\nabla_\theta\log\pi_\theta(\sigma_t^+\mid\tilde h_t)
+\sum_{i,j}\nabla_\theta d_{i,j}(\theta)
\right].
\label{eq:app_post_exact}
\end{align}
Here, the continuation distribution includes environment transitions, which are independent of $\theta$.
The first term accounts for how the policy changes the distribution of visited prefixes; the second differentiates the token-level KL at fixed prefixes.

Following ~\citet[Appendix~A.1]{hubotter2026reinforcement}, we use per-token estimator and apply stop-gradient through continuation sampling process.
Specifically, we use the surrogate
\begin{equation}
\overline D_t(m_t)
= \mathbb{E}_{\sigma_t^+\sim\mathrm{sg}[\pi_\theta(\cdot\mid\tilde h_t)]}
\!\left[\sum_{i,j}d_{i,j}(\theta)\right],
\qquad
g_{\mathrm{post}}^{\mathrm{SD}}
= -\mathbb{E}_{m_t}\!\left[\nabla_\theta\overline D_t(m_t)\right].
\label{eq:app_post_surrogate}
\end{equation}
This gives a gradient of the surrogate rather than the full continuation objective.
The continuation is still sampled from the current policy but the sampling process is detached during differentiation.

To differentiate each KL, abbreviate the student and stopped teacher distributions at a fixed prefix as
$p_\theta(v)=\pi_\theta(v\mid\tilde h_t,s_{i,j})$ and
$q(v)=\mathrm{sg}[\pi_\theta(v\mid h_t,s_{i,j})]$.
Then
\begin{align}
-\nabla_\theta d_{i,j}(\theta)
&= -\nabla_\theta\sum_v p_\theta(v)\log\frac{p_\theta(v)}{q(v)}
\nonumber\\
&= -\sum_v\nabla_\theta p_\theta(v)
\left(\log\frac{p_\theta(v)}{q(v)}+1\right)
\nonumber\\
&= \mathbb{E}_{v\sim p_\theta}\!\left[
\log\frac{q(v)}{p_\theta(v)}\nabla_\theta\log p_\theta(v)
\right],
\label{eq:app_token_kl_gradient}
\end{align}
where the last equality uses $\sum_v\nabla_\theta p_\theta(v)=\nabla_\theta 1=0$.
Using the sampled action token $a_{i,j}$ for this inner expectation yields
\begin{equation}
g_{\mathrm{post}}^{\mathrm{SD}}
= \mathbb{E}_{m_t,\sigma_t^+}\!\left[
\sum_{i,j}
\log\frac{\pi_\theta(a_{i,j}\mid h_t,s_{i,j})}
{\pi_\theta(a_{i,j}\mid\tilde h_t,s_{i,j})}
\nabla_\theta\log\pi_\theta(a_{i,j}\mid\tilde h_t,s_{i,j})
\right].
\label{eq:app_post_sd}
\end{equation}
A sampled continuation therefore provides a Monte Carlo estimate of this surrogate gradient, with the log-ratio treated as a fixed token-level advantage.
Equivalently, this update can be interpreted by using discount factor $\gamma=0$ for the distillation reward, where each token receives its own teacher--student log-ratio, without rewards from later tokens~\citep{lu2025onpolicydistillation}.

\newpage

\section{Implementation of \method}

\subsection{Training Algorithm}
\label{app:algorithm_implementation}

Figure~\ref{fig:recap_pseudocode} summarizes the computation of the \method{} loss.

\definecolor{recapCodeBackground}{HTML}{F2F2F2}
\definecolor{recapCodeKeyword}{HTML}{2B7D16}
\definecolor{recapCodeComment}{HTML}{497F87}
\definecolor{recapCodeString}{HTML}{C72828}

\begin{figure}[!ht]
\hrule height 0.7pt
\begin{lstlisting}[
    language=Python,
    basicstyle=\ttfamily\footnotesize\color{black},
    keywordstyle=\bfseries\color{recapCodeKeyword},
    commentstyle=\color{recapCodeComment},
    stringstyle=\color{recapCodeString},
    emph={recap_loss}, emphstyle=\color{blue},
    numbers=left, numberstyle=\scriptsize\rmfamily\color{black},
    numbersep=8pt, stepnumber=1, numberblanklines=true,
    columns=fullflexible, keepspaces=true, showstringspaces=false,
    frame=none, backgroundcolor={},
    xleftmargin=24pt, xrightmargin=8pt, aboveskip=7pt, belowskip=7pt
]
def recap_loss(batch, weight):
    # Cached rollout scores on the sampled tokens.
    old_logp = batch.rollout_logprobs
    comp_adv = zeros_like(old_logp)
    rewards = []
    for event in batch.compactions:
        idx = event.continuation_tokens
        sampled_kl = old_logp[idx] - batch.teacher_logprobs[idx]
        rewards.append(-sampled_kl.mean())
        comp_adv[idx] = -sampled_kl / len(idx)

    # Leave-one-out baseline within each task and compaction index.
    summary_adv = leave_one_out(rewards, batch.compaction_groups)
    comp_adv[batch.summary_tokens] = broadcast(summary_adv)

    # Combine advantages before the shared clipped policy loss.
    task_adv = grpo(batch.rewards, batch.task_groups)
    advantage = (task_adv + weight * comp_adv).detach()
    logp = pi_theta.log_prob(batch)
    ratio = exp(logp - old_logp)
    loss = clipped_policy_loss(ratio, advantage)
    return aggregate(loss, batch, "trajectory-mean-token-mean")
\end{lstlisting}
\hrule height 0.7pt
\caption{Pseudocode for \method.}
\label{fig:recap_pseudocode}
\end{figure}

\noindent\textbf{Implementation details.}
We estimate $A^{\mathrm{task}}$ with GRPO~\citep{shao2024deepseekmath} by subtracting the task-group mean from each terminal reward and dividing by the group's reward standard deviation.
This advantage is shared by all action and summary tokens in the trajectory.
For $g_{\mathrm{sum}}$, we estimate the baseline $b_t$ using the leave-one-out mean~\citep{ahmadian2024back} of negative distortion rewards from other rollouts of the same task at the same compaction index.
We average token losses within each complete trajectory, including all its segments, and then average over trajectories.
We reuse rollout-policy $\pi_\mathrm{rollout}$ for the calculation in both the importance-sampling ratio $\frac{\pi_\theta}{\pi_{\mathrm{rollout}}}$ and in the KL estimate between compressed and original contexts.
For each token-level KL, we use the log-probability difference at the sampled token instead of the approximation with the top-$k$ tokens in the vocabulary, as we did not observe a clear benefit from the latter in our experiments.

\subsection{Extension to Multiple Compactions}
\label{app:multiple_compaction}

For multiple compactions, each event assigns advantages to its own summary and the following actions up to the next compaction or task termination, so the affected tokens do not overlap across events. We compute these advantages separately, combine them with the task advantages, and aggregate the resulting token losses together in the trajectory level. 
We always treat pre-compaction history $h_t$ as fixed for each compaction event. Thus, distortion at later compactions does not provide a learning signal to earlier compactions, even though later histories contain earlier summaries; for example, $h_t$ at the second compaction includes the first summary.

\newpage

\section{Experiment Details}
\label{app:experiment_details}

For each environment, we select 300 training problems of suitable difficulty using preliminary Qwen3.5-4B rollouts, favoring problems with mixed outcomes while dropping too easy questions, following~\citet{he2026agentlightningv10harnessed} and \citet{kim2026frognano}.
We use BrowseComp-Plus~\citep{chen2025browsecomp} for deep search and SWE-smith~\citep{yang2026swe} for software engineering, evaluating on 150 held-out BrowseComp-Plus questions and all 500 SWE-bench Verified tasks~\citep{jimenez2024swe}.
All methods share the same training subsets and environment settings.

\noindent\textbf{BrowseComp-Plus.}
We use Qwen3-Embedding-8B for dense retrieval over the fixed BrowseComp-Plus corpus.
The \texttt{search} tool returns the top five documents with their IDs, retrieval scores, and first 512 tokens; \texttt{get\_document} retrieves a document by ID and returns its first 4,096 tokens.
We use Qwen3.8-27B as the LLM-judge for checking semantic equivalence for the remaining answers, assigning binary reward to the agent's answer.

\noindent\textbf{SWE-bench Verified.}
Each task runs in an isolated sandbox with \texttt{bash} for commands and tests, \texttt{str\_replace\_editor} for file inspection and editing, and \texttt{finish} for patch submission.
File changes persist across turns; one tool call is allowed per turn.
Truncated shell output retains its first and last 3,000 characters; editor output is capped at 20,000 characters.
Patches are evaluated in a fresh sandbox using task-specific tests and, for SWE-bench Verified, the official grader.

\textbf{Selection of $\lambda$.}
We select $\lambda$ separately for each environment using a held-out validation set of 100 problems.
We compare $\lambda\in\{0.03,0.1,0.3,1.0\}$ and choose the value with the highest validation accuracy (Table~\ref{tab:app_lambda_selection}), yielding $\lambda=0.1$ for BrowseComp-Plus and $\lambda=0.03$ for software engineering.

\begin{table}[!ht]
    \centering
    \caption{\textbf{Selection of $\lambda$.} We evaluated \method with different choices of $\lambda$ on the held out validation set and select the best for each environment.}
    \label{tab:app_lambda_selection}
    \small
    \setlength{\tabcolsep}{9pt}
    \renewcommand{\arraystretch}{1.1}
    \vspace{1em}
    \begin{tabular}{@{}ccccc@{}}
        \toprule
        & \multicolumn{2}{c}{\textbf{BrowseComp-Plus}}
        & \multicolumn{2}{c}{\textbf{SWE-agent}} \\
        \cmidrule(lr){2-3}\cmidrule(l){4-5}
        $\lambda$ & Val. & Test & Val. & Test \\
        \midrule
        0.03 & 35.0 & 60.0 & \textbf{44.0} & 51.1 \\
        0.1 & \textbf{40.0} & 62.8 & 39.0 & 46.3 \\
        0.3 & 36.0 & 62.3 & 41.0 & 46.5 \\
        1.0 & 34.0 & 62.3 & 43.0 & 49.8 \\
        \bottomrule
    \end{tabular}
\end{table}

\newpage
\textbf{Training Details}. 
We train Qwen3.5-4B with thinking disabled using fully asynchronous VeRL and the hyperparameters in Table~\ref{tab:app_training_hyperparameters}.
Evaluation uses temperature $1.0$, top-$p=0.95$, and top-$k=20$, with accuracy averaged over four runs.
All experiments run on NVIDIA GH200 GPUs.

\begin{table}[!ht]
    \centering
    \caption{Training configurations of \method shared across environments.}
    \label{tab:app_training_hyperparameters}
    \small
    \setlength{\tabcolsep}{6pt}
    \renewcommand{\arraystretch}{1.15}
    \vspace{1em}
    \begin{tabular}{@{}lc@{}}
        \toprule
        \textbf{Setting} & \textbf{Value} \\
        \midrule
        Maximum interaction steps (BrowseComp / SWE) & 100 / 200 \\
        Maximum output tokens per response & 8,192 \\
        Task groups per training step & 16 \\
        Rollouts per task (group size) & 8 \\
        Mini-batch size (task groups) & 8 \\
        Inference engine & vllm \\
        Training temperature & $1.0$ \\
        Training steps & 150 \\
        \midrule
        Optimizer & AdamW \\
        Learning rate & $1\times10^{-6}$ \\
        Weight decay & $0.01$ \\
        Warmup steps & 0 \\
        Gradient clipping norm & $1.0$ \\
        PPO clipping $(\epsilon_{\mathrm{low}},\epsilon_{\mathrm{high}})$ & $(0.20,0.28)$ \\
        Entropy & $0$ \\ 
        Reference-policy KL & $0$ \\ 
        \method{} weight $\lambda$ (BrowseComp-Plus / SWE)  & 0.1 / 0.03 \\
        \bottomrule
    \end{tabular}
\end{table}

\newpage

\section{Prompts}
\label{app:prompts}

\tcbset{recapprompt/.style={
  colback=black!4, colframe=black, coltitle=white,
  boxrule=0.65pt, arc=1.5mm, boxsep=0pt,
  left=8pt, right=8pt, top=6pt, bottom=6pt,
  toptitle=3pt, bottomtitle=3pt,
  fonttitle=\normalsize\rmfamily,
  fontupper=\normalsize\rmfamily,
  before skip=8pt, after skip=8pt
}}

\lstdefinestyle{recapschema}{
  basicstyle=\fontsize{8.5}{10}\selectfont\ttfamily,
  columns=fullflexible, keepspaces=true,
  breaklines=true, breakatwhitespace=true, breakindent=2em,
  showstringspaces=false, aboveskip=0pt, belowskip=0pt
}

\subsection{Context Compaction}
\label{app:prompt_compaction}

We use one shared prompt of context compaction for both environments.
At each compaction, we append the following instruction to the current conversation.

\begin{tcolorbox}[recapprompt,title={Context Compaction Prompt}]
\raggedright
\setlength{\parindent}{0pt}
\setlength{\parskip}{0.45em}
The current task is paused for context compaction. For this turn, write a handoff summary for another agent that will resume the original task.

Include:

- Current progress and key decisions made\\
- Important context, constraints, or observations\\
- What remains to be done (clear next steps)\\
- Any critical data, examples, or references needed to continue\\
- Be concise, structured, and focused on helping the next agent seamlessly continue the work.

Do not call any tools in this turn. Return only the handoff summary. The entire response will be used as the summary for the next agent.
\end{tcolorbox}

After compaction, we keep the original task prompt and add the instruction and summary below as a new user message.
We also append the last $k$ iterations, each consisting of an agent action and an environment observation.
We use $k=1$ in our experiments.
The previous summary and any interactions before these $k$ iterations are removed in the new context.

\begin{tcolorbox}[recapprompt,title={Context After Compaction}]
\raggedright
\setlength{\parindent}{0pt}
\setlength{\parskip}{0.45em}
\{original\_task\_prompt\}

Another agent started to solve this problem and produced a summary of its thinking process. You also have access to the actions and observations that were obtained by that agent. Use this to build on the work that has already been done and avoid duplicating work. Here is the summary produced by the other agent, use the information in this summary to assist with your own analysis:

\{summary\}

\{previous $k$ iterations\}
\end{tcolorbox}

\clearpage
\subsection{BrowseComp-Plus}

\begin{tcolorbox}[recapprompt,title={BrowseComp-Plus Prompt}]
\raggedright
\setlength{\parindent}{0pt}
\setlength{\parskip}{0.45em}
You are a deep research agent. You need to answer the given question by interacting with a search engine, using the \textit{search} and \textit{get\_document} tools provided. Please perform reasoning and use the tools step by step, in an interleaved manner. You may use the search and get\_document tools multiple times.

NOTE:\\
- You should always call one tool at a time. Use a concise, targeted query containing the most discriminative clues for each tool call.\\
- Before each tool call, provide your reasoning process to explain what you have learned and why this is the next useful call.\\
- When you have found the answer, call the \textit{finish} tool with your succinct final answer.

Question: \{query\}
\end{tcolorbox}
\begin{tcblisting}{recapprompt,listing only,breakable,
  title={BrowseComp-Plus Tool Schemas},title after break={BrowseComp-Plus Tool Schemas (continued)},
  listing options={style=recapschema}}
{
  "name": "search",
  "description": "Perform a search on a knowledge source. Returns top-5 hits with docid, score, and snippet. The snippet contains the document's contents and may be truncated.",
  "parameters": {
    "type": "object",
    "properties": {
      "query": {"type": "string", "description": "Search query string."}
    },
    "required": ["query"]
  }
}

{
  "name": "get_document",
  "description": "Retrieve the first 4,096 tokens of a document by its docid.",
  "parameters": {
    "type": "object",
    "properties": {
      "docid": {"type": "string", "description": "Document ID to retrieve."}
    },
    "required": ["docid"]
  }
}

{
  "name": "finish",
  "description": "Submit the succinct final answer and end the episode.",
  "parameters": {
    "type": "object",
    "properties": {
      "answer": {"type": "string", "description": "Succinct final answer."}
    },
    "required": ["answer"]
  }
}
\end{tcblisting}

% Keep the environment heading with its complete prompt.
\clearpage
\subsection{Software Engineering}
\label{app:prompt_swe_agent}

\begin{tcolorbox}[recapprompt,title={Software Engineering Task Prompt}]
\raggedright
\setlength{\parindent}{0pt}
\setlength{\parskip}{0.45em}
You are an autonomous software engineer working in the repository at /testbed. Your goal is to resolve the issue described by the user with a correct and maintainable patch.

\#\# Recommended workflow

Work step by step so you can learn from command output and test results:

1. Explore the repository and read the relevant code.\\
2. Reproduce or clearly identify the reported problem.\\
3. Determine the underlying cause.\\
4. Edit the source code to fix the issue.\\
5. Run focused tests or a reproduction to verify the fix.\\
6. Check relevant edge cases and run broader tests when practical.\\
7. Submit the completed work with the finish tool.

Adapt this workflow when a step is unnecessary or impractical.

\#\# Tool-use rules

While actively solving the issue, briefly explain what you learned and why the next action is useful, then call one tool: \textit{bash, str\_replace\_editor, or finish}. Do not make multiple or parallel tool calls in one response.

Use \textit{bash} to inspect the repository, search code, run programs, and execute tests. Use \textit{str\_replace\_editor} for precise file viewing, creation, and editing.

The repository filesystem persists across tool calls. Each bash call starts in /testbed, but shell directory changes and environment variables do not persist to later calls.

Avoid commands that produce unnecessarily large output. Prefer targeted file views, searches, and test commands. If a command fails, inspect its output and continue diagnosing the problem.

When the issue is fixed and adequately tested, briefly summarize the verification and call \textit{finish} with empty arguments. After calling finish, you cannot make further changes.

\#\# Issue

\{problem\_statement\}
\end{tcolorbox}

\clearpage
The available tools are \texttt{bash(command)}, \texttt{str\_replace\_editor(command, path)} and \texttt{finish()} for SWE-agent.

\begin{tcblisting}{recapprompt,listing only,breakable,
  title={Software Engineering Tool Schemas},title after break={Software Engineering Tool Schemas (continued)},
  listing options={style=recapschema}}
{
  "name": "bash",
  "description": "Run one shell command in the repository. Each call starts a fresh bash shell in /testbed, while filesystem changes persist across calls.",
  "parameters": {
    "type": "object",
    "properties": {
      "command": {"type": "string", "description": "The bash command to run in /testbed."}
    },
    "required": ["command"],
    "additionalProperties": false
  }
}

{
  "name": "str_replace_editor",
  "description": "Custom editing tool for viewing, creating, and editing plain-text files. State persists for this task. File views use cat -n style line numbers; directory views list sorted non-hidden entries up to two levels. str_replace requires old_str to match exactly once.",
  "parameters": {
    "type": "object",
    "properties": {
      "command": {"type": "string", "enum": ["view", "create", "str_replace", "insert"], "description": "The editor operation to perform."},
      "path": {"type": "string", "description": "Absolute path below /testbed, or a relative path from /testbed (e.g. src/file.py)."},
      "file_text": {"type": "string", "description": "Complete file content required by create."},
      "old_str": {"type": "string", "description": "Exact unique text required by str_replace."},
      "new_str": {"type": "string", "description": "Replacement text for str_replace, or text required by insert. Omit it from str_replace to delete old_str."},
      "insert_line": {"type": "integer", "description": "Insert new_str after this one-based line. Use 0 to insert before the first line."},
      "view_range": {"type": "array", "items": {"type": "integer"}, "minItems": 2, "maxItems": 2, "description": "Optional inclusive [start_line, end_line] for file view. Use -1 as end_line to read through the end."}
    },
    "required": ["command", "path"],
    "additionalProperties": false
  }
}

{
  "name": "finish",
  "description": "Finish the task and submit the current repository changes for evaluation.",
  "parameters": {
    "type": "object",
    "properties": {},
    "additionalProperties": false
  }
}
\end{tcblisting}

\end{document}